\documentclass[conference]{IEEEtran}
\IEEEoverridecommandlockouts
\usepackage{cite}
\usepackage{amsmath,amssymb,amsfonts}
\usepackage{algorithmic}
\usepackage{graphicx}
\usepackage{textcomp}
\usepackage{xcolor}
\def\BibTeX{{\rm B\kern-.05em{\sc i\kern-.025em b}\kern-.08em
    T\kern-.1667em\lower.7ex\hbox{E}\kern-.125emX}}

\usepackage{subfigure}
\usepackage[pagebackref=true,breaklinks=true,colorlinks,bookmarks=false]{hyperref}
\usepackage[nameinlink]{cleveref}
    
\DeclareMathOperator*{\argmax}{arg\, max}
\newcommand{\IoU}{\mathit{IoU}}

\newcommand{\sIoU}{\mathit{IoU}_{\mathrm{adj}}}

\definecolor{slateblue}{rgb}{0.42, 0.35, 0.8}
\definecolor{indianred}{rgb}{0.8, 0.36, 0.36}
\definecolor{mediumseagreen}{rgb}{0.24, 0.7, 0.44}
\definecolor{darkgoldenrod}{rgb}{0.72, 0.53, 0.04}
\definecolor{mediumred-violet}{rgb}{0.73, 0.2, 0.52}
\definecolor{cornflowerblue}{rgb}{0.39, 0.58, 0.93}
\definecolor{goldenyellow}{rgb}{1.0, 0.87, 0.0}
\definecolor{darkseagreen}{rgb}{0.56, 0.74, 0.56}
\definecolor{lightskyblue}{rgb}{0.53, 0.81, 0.98}
\definecolor{mediumslateblue}{rgb}{0.48, 0.41, 0.93}
\definecolor{plum(web)}{rgb}{0.8, 0.6, 0.8}
\definecolor{mediumorchid}{rgb}{0.73, 0.33, 0.83}

\begin{document}

\title{Time-Dynamic Estimates of the Reliability of Deep Semantic Segmentation Networks}

\author{\IEEEauthorblockN{Kira Maag}
\IEEEauthorblockA{\textit{Department of Mathematics} \\
\textit{University of Wuppertal, Germany}\\
kmaag@uni-wuppertal.de}
\and
\IEEEauthorblockN{Matthias Rottmann}
\IEEEauthorblockA{\textit{Department of Mathematics} \\
\textit{University of Wuppertal, Germany}\\
rottmann@uni-wuppertal.de}
\and
\IEEEauthorblockN{Hanno Gottschalk}
\IEEEauthorblockA{\textit{Department of Mathematics} \\
\textit{University of Wuppertal, Germany}\\
hgottsch@uni-wuppertal.de}
}

\maketitle

\begin{abstract}
In the semantic segmentation of street scenes with neural networks, the reliability of predictions is of highest interest. The assessment of neural networks by means of uncertainties is a common ansatz to prevent safety issues. As in applications like automated driving, video streams of images are available, we present a time-dynamic approach to investigating uncertainties and assessing the prediction quality of neural networks. We track segments over time and gather aggregated metrics per segment, thus obtaining time series of metrics from which we assess prediction quality. This is done by either classifying between intersection over union equal to 0 and greater than 0 or predicting the intersection over union directly. We study different models for these two tasks and analyze the influence of the time series length on the predictive power of our metrics.
\end{abstract}


\section{Introduction}
\begin{figure}[t]
    \center
    \includegraphics[width=0.32\textwidth]{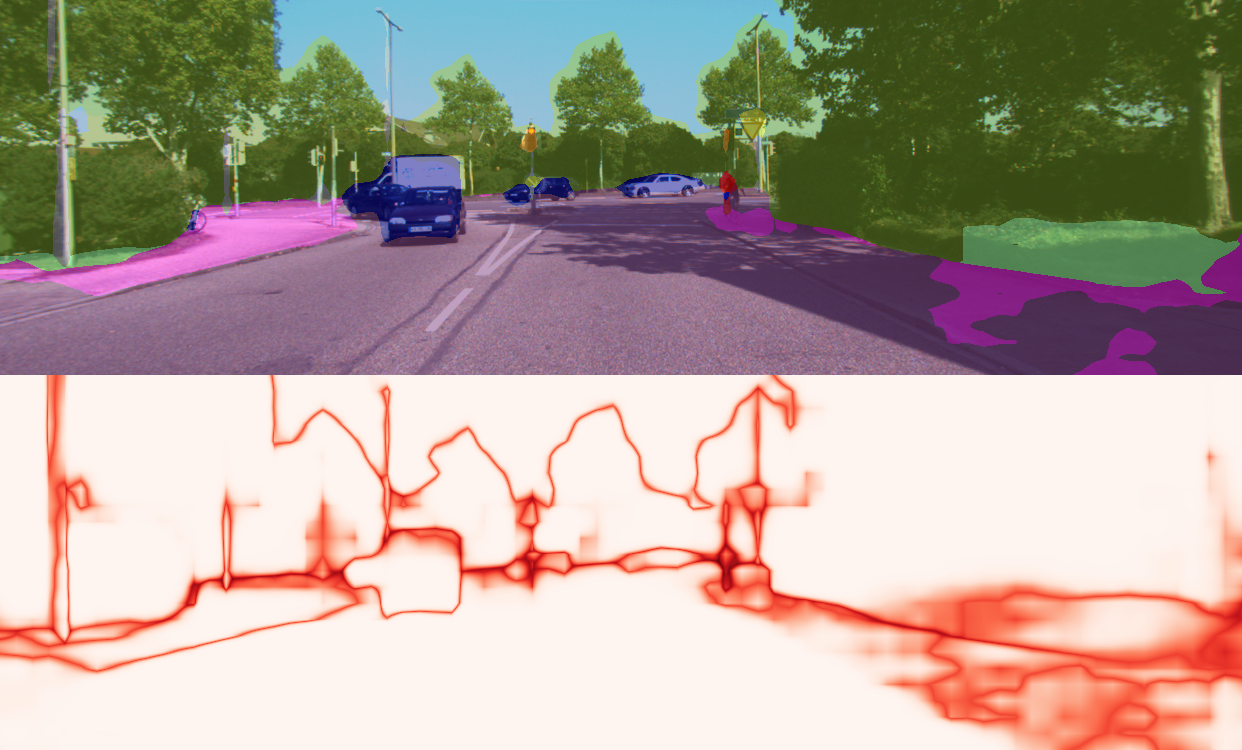}
    \caption{Segmentation predicted by a neural network (\emph{top}) and variation ratio (\emph{bottom}).}
    \label{fig:metrics}
\end{figure}

Semantic segmentation, i.e., the pixel-wise classification of image content, is an important tool for scene understanding. In recent years, neural networks have demonstrated outstanding performance for this task. In safety relevant applications like automated driving \cite{Huang2018} and medical imaging \cite{Wickstrom2018}, the reliability of predictions and thus uncertainty quantification is of highest interest. While most works focus on uncertainty quantification for single-frames, there is often video data available. In this work, we investigate uncertainties by taking temporal information into account. To this end, we construct metrics that express uncertainties in single frames. By tracking predicted segments over time, we obtain time series of metrics that quantify the dynamics of predicted objects. From this information we assess the prediction quality on segment-level. 
\paragraph{Uncertainty Quantification.} 
A very important type of uncertainty is the model uncertainty resulting from the fact that the ideal parameters are unknown and have to be estimated from data. Bayesian models are one possibility to consider these uncertainties \cite{Mackay1992}. Therefore, different frameworks based on variational approximations for Bayesian inference exist \cite{Attias2000,Duvenaud2016}.
Recently, Monte-Carlo (MC) Dropout \cite{Gal2016} as approximation to Bayesian inference has aroused a lot of interest. In classification tasks, the uncertainty score can be directly determined on the network's output \cite{Gal2016}. Threshold values for the highest softmax probability or threshold values for the entropy of the classification distributions (softmax output) are common approaches for the detection of false predictions (false positives) of neural networks, see e.g.~\cite{Liang2017}.
Uncertainty metrics like classification entropy or the highest softmax probability are usually combined with model uncertainty (MC Dropout inference) or input uncertainty, see \cite{Gal2016} and \cite{Liang2017}, respectively. Alternatively, gradient-based uncertainty metrics are proposed in \cite{Oberdiek2018} and an alternative to Bayesian neural networks is introduced in \cite{Lakshminarayanan2017} where the idea of ensemble learning is used to consider uncertainties.
These uncertainty measures have proven to be practically efficient for detecting uncertainty and some of them have also been transferred to semantic segmentation tasks, such as MC Dropout, which also achieves performance improvements in terms of segmentation accuracy, see~\cite{Kendall2015}.
The work presented in \cite{Wickstrom2018} also makes use of MC Dropout to model uncertainty and filter out predictions with low reliability. This line of research is further developed in \cite{Huang2018} to detect spatial and temporal uncertainty in the semantic segmentation of videos. In semantic segmentation tasks the concepts of \emph{meta classification} and \emph{meta regression} are introduced in \cite{Rottmann2018}. Meta classification refers to the task of predicting whether a predicted segment intersects with the ground truth or not. 
Therefore, the intersection over union ($\IoU$, also known as Jaccard index \cite{Jaccard1912}), a commonly used performance measure for semantic segmentation, is considered. The $\IoU$ quantifies the degree of overlap of prediction and ground truth, it is equal to zero if and only if the predicted segment does not intersect with the ground truth. The meta classification task corresponds to (meta) classifying between $\IoU=0$ and $\IoU>0$ for every predicted segment. Meta regression is the task of predicting the $\IoU$ 
for each predicted segment directly. The main aim of both tasks is to have a model that is able to reliably assess the quality of a semantic segmentation obtained from a neural network. The predicted $\IoU$ therefore also serves as a performance estimate. As input both methods use segment-wise metrics extracted from the segmentation network's softmax output. 
The same tasks are pursued in \cite{DeVries2018,Huang2016} for images containing only a single object, instead of metrics they utilize additional CNNs. 
In \cite{Schubert2019} the work of \cite{Rottmann2018} is extended by adding resolution dependent uncertainty and further metrics. In \cite{Erdem2004} performance measures for the segmentation of videos are introduced, these measures are also based on image statistics and can be calculated without ground truth.
\paragraph{Visual Object Tracking.} 
Object tracking is an essential task in video applications, such as automated driving, robot navigation and many others. The tasks of object tracking consist of detecting objects and then tracking them in consecutive frames, eventually studying their behavior \cite{Yilmaz2006}. In most works, the target object is represented as an axis-aligned or rotated bounding box \cite{Kristan2015}.
The following approaches work with bounding boxes. A popular strategy for object tracking is the tracking-by-detection approach \cite{Babenko2009}. A discriminative classifier is trained online while performing the tracking to separate the object from the background only by means of the information where the object is located in the first frame. Another approach for tracking-by-detection uses adaptive correlation filters that model the targets appearance, the tracking is then performed via convolution with the filters \cite{Bolme2010}. In \cite{Danelljan2015} and \cite{Valmadre2017}, the trackers based on correlation filters are improved with spatial constraints and deep features, respectively.
Another object tracking algorithm \cite{Mu2016} combines Kalman filters and adaptive least squares to predict occluded objects where the detector shows deficits. In contrast to online learning, there are also tracking algorithms that learn the tracking task offline and perform tracking as inference, only. 
The idea behind these approaches is to train offline a similarity function on pairs of video frames instead of training a discriminative classifier online \cite{Bertinetto2016}.
In \cite{Bertinetto2016} a fully-convolutional siamese network is introduced. This approach is improved by making use of region proposals \cite{Li2018}, angle estimation and spatial masking \cite{He2018} as well as memory networks \cite{Yang2018}. Another approach for object tracking with bounding boxes is presented in \cite{Yao2019} where semantic information is used for tracking. Most algorithms and also the ones described here use bounding boxes, mostly for initializing and predicting the position of an object in the subsequent frames. In contrast, \cite{Comaniciu2000} uses coarse binary masks of target objects instead of rectangles. There are other procedures that initialize and/or track an object without bounding boxes, since a rectangular box does not necessarily capture the shape of every object well. 
Other approaches use semantic image segmentation such as \cite{Hariharakrishnan2005}, where the initialization includes a segmentation for predicting object boundaries. Segmentation-based tracking algorithms are presented in \cite{Aeschliman2010} and \cite{Duffner2013} based on a pixel-level probability model and an adaptive model, respectively. 
The approaches presented in \cite{Belagiannis2012} and \cite{Son2015} are also based on segmentation and use particle filters for the tracking process. There are also superpixel-based approaches, see e.g.~\cite{Yeo2017}, and a fully-convolutional siamese approach \cite{Wang2018} that creates binary masks and starts from a bounding box initialization.
\paragraph{Our Contribution.} 
In this work we elaborate on the meta classification and regression approach from \cite{Rottmann2018} that provides a framework for post processing a semantic segmentation. This method generates uncertainty heat maps from the softmax output of the semantic segmentation network, such as pixel-wise entropy, probability margin or variation ratio (see \cref{fig:metrics}). 
In addition to these segment-wise metrics, further quantities derived from the predicted segments are used, for instance various measures corresponding to the segments geometry. This set of metrics, yielding a structured dataset where each row corresponds to a predicted segment, is presented to meta classifier/regressor to either classify between $\IoU=0$ and $\IoU>0$ or predict the $\IoU$ directly. In contrast to \cite{Rottmann2018} we use the additional metrics proposed in \cite{Schubert2019}. In this paper, we extend the work presented in \cite{Rottmann2018} by taking time-dynamics into account. 
To the best of our knowledge this is the first work that detects false positive segments with the help of time-dynamic metrics.
A core assumption is that a semantic segmentation network and a video stream of input data are available. We present a light-weight approach for tracking semantic segments over time, matching them according to their overlap in consecutive frames. This is leveraged by shifting segments according to their expected location in the subsequent frame. The obtained time series of metrics are presented as input to meta classifiers and regressors. For the latter we study different types of models and their dependence on the length of the time series. 

In our tests, we employ two publicly available DeepLabv3+ networks \cite{Chen2018} and apply them to the VIsual PERception (VIPER) dataset \cite{Richter2017} and to the KITTI dataset \cite{Geiger2013}. For the synthetic VIPER dataset we train a DeepLabv3+ network and demonstrate that the additional information from our time-dynamic approach improves over its single-frame counterpart \cite{Rottmann2018} w.r.t.\ both meta classification and regression (meta tasks). 
Furthermore, the different models used for the meta tasks yield additional improvement.
For meta regression we obtain $R^2$ values of up to $85.82\%$ and for meta classification AUROC values of up to $86.01\%$.
For the VIPER dataset there are labeled ground truth images for each frame, while for the KITTI dataset only a few frames per video are labeled with ground truth. Hence, for the KITTI datset we use alternative sources of useful information besides the real ground truth, i.e., pseudo ground truth provided by a stronger network.
For meta regression we achieve $R^2$ values of up to $87.51\%$ and for the meta classification AUROC values of up to $88.68\%$, also outperforming \cite{Rottmann2018}.
Given the fact that for automated driving video sequences are available, the proposed light-weight tracking algorithm uses information from consecutive frames to leverage meta classification and regression performance with only low additional effort.
\paragraph{Related Work.} 
Most works 
\cite{Kristan2015,Babenko2009,Bolme2010,Danelljan2015,Valmadre2017,Mu2016,Xiang2015,Bertinetto2016,Li2018,He2018,Yang2018,Yao2019} in the field of object tracking make use of bounding boxes while our approach is based on semantic segmentation. There are some approaches that make use of segmentation masks. However, only a coarse binary mask is used in \cite{Comaniciu2000} and in \cite{Hariharakrishnan2005} the segmentation is only used for initialization. In \cite{Aeschliman2010,Duffner2013} segmentation and tracking are executed jointly.
In our procedure, a segmentation is inferred first, tracking is performed afterwards. In addition to the different forms of object representations, there are various algorithms for object tracking. In the tracking-by-detection methods a classifier for the difference between object and background is trained and therefore only information about the location of the object in the first frame is given \cite{Babenko2009,Bolme2010,Danelljan2015,Valmadre2017}. 
We do not train classifiers as this information is contained in the inferred segmentations. Another approach is to learn a similarity function offline \cite{Bertinetto2016,Li2018,He2018,Yang2018}. 
The works of \cite{Aeschliman2010,Duffner2013,Belagiannis2012,Son2015,Wang2018} are based on segmentation and they use different tracking methods, like probability models, particle filters and fully-convolutional siamese network, respectively. Our tracking method is solely based on the degree of overlap of predicted segments.

With respect to uncertainty quantification, MC dropout is widely used, see 
\cite{Gal2016,Kendall2015,Wickstrom2018}. Whenever dropout is used in a segmentation network (we do not use dropout), the resulting heat map can be equipped by our framework. 
There are alternative measures of uncertainty like gradient based ones \cite{Oberdiek2018} or measures based on spatial and temporal differences between the colors and movements of the objects \cite{Erdem2004}. We construct metrics based on aggregated dispersion measures from the softmax output of a neural network at segment level. 
The works \cite{DeVries2018,Huang2016} closest to ours are constructed to work with one object per image, instead of hand crafted metrics they are based on post-processing CNNs.
We extend the work of \cite{Rottmann2018} by a temporal component, and further investigate methods for the meta classification and regression, e.g.\ gradient boosting and neural networks. 
\paragraph{Outline.} 
The remainder of this work is organized as follows. In \cref{sec:tracking_segments} we introduce a tracking algorithm for semantic segmentation. This is followed by the construction of segment-wise metrics using uncertainty and geometry information in \cref{sec:metrics}. In \cref{sec:pred} we describe the meta regression and classification methods including the construction of their inputs consisting of time series of metrics. Finally, we present numerical results in \cref{sec:result}. We study the influence of time-dynamics on meta classification and regression as well as the incorporation of various classification and regression methods.
%
%
%
\section{Tracking Segments over Time}\label{sec:tracking_segments}
In this section we introduce a light-weight tracking method for the case where a semantic segmentation is available for each frame of a video. Semantic image segmentation aims at segmenting objects in an image. It can be viewed as a pixel-wise classification of image content (e.g.~top panel of \cref{fig:metrics}). 
To obtain a semantic segmentation, the goal is to assign to each image pixel $z$ of an input image $x$ a label $y$ within a prescribed label space $\mathcal{C} = \{y_{1}, \ldots, y_{c} \}$. 
Here, this task is performed by a neural network that provides for each pixel $z$ a probability distribution $f_{z}(y|x,w)$ over the class labels $y \in \mathcal{C}$, given learned weights $w$ and an input image $x$. The predicted class for each pixel $z$ is obtained by
\begin{equation}
    \hat y_z(x,w)=\argmax_{y\in\mathcal{C}}f_z(y|x,w) \, .
\end{equation}
Let $\hat{\mathcal S}_x=\{\hat y_z(x,w) | z\in x\}$ denote the predicted segmentation and $\hat{\mathcal K}_x$ the set of predicted segments. A segment is defined as a connected component of which all pixels belong to the same class. The idea of the proposed tracking method is to match segments of the same class according to their overlap in consecutive frames. We denote by $\{ x_{1}, \ldots, x_{T} \}$ the image sequence with a length of $T$ and $x_{t}$ corresponds to the $t^\mathit{th}$ image. Furthermore, we formulate the overlap of a segment $k$ with a segment $j$ through
\begin{equation}
    O_{j,k} = \frac{| j \cap k |}{| j |} \, .
\end{equation}
To account for motion of objects, we also register geometric centers of predicted segments. The geometric center of a segment $k \in \hat{\mathcal K}_{x_{t}}$ in frame $t$ is defined as
\begin{equation}\label{geom_center}
    \bar{k}_{t} = \frac{1}{|k|} \sum_{z \in k} z
\end{equation}
where $z=(z_{1}, z_{2})$ is given by its vertical and horizontal coordinates of pixel $z$.

Our tracking algorithm is applied sequentially to each frame $t$, $t = 1, \ldots, T$, and we aim at tracking all segments present in at least one of the frames. To give the segments different priorities for matching, the segments of each frame are sorted by size and treated in descending order. As is the case when a segment in frame $t$ has been matched with a segment from previous frames, it is ignored in further steps and matched segments are assigned an id. 
Within the description of the matching procedure, we introduce parameters $c_\mathit{near}$, $c_\mathit{over}$, $c_\mathit{dist}$ and $c_\mathit{lin}$, the respective numerical choices are given in \cref{sec:result}. More formally, our algorithm consists of the following five steps: 

\textbf{Step 1 (aggregation of segments).} $ \ $ The minimum distance between segment $i \in \hat{\mathcal K}_{x_{t}}$ and all $j \in \hat{\mathcal K}_{x_{t}} \setminus \{i\}$ of the same class is calculated. If the distance is less than a constant $c_\mathit{near}$, the segments are so close to each other that they are regarded as one segment and receive a common id. 

\textbf{Step 2 (shift).} $ \ $ If the algorithm was applied to at least two previous frames, the geometric centers $\left(\bar{k}_{t-2} \right)$ and $\left(\bar{k}_{t-1} \right)$ of segment $k \in \hat{\mathcal K}_{x_{t-1}}$ are computed. The segment from frame $t-1$ is shifted by the vector $ \left( \bar{k}_{t-1} - \bar{k}_{t-2} \right)$ and the overlap $O_{j,k}$ with each segment $j \in \hat{\mathcal K}_{x_{t}}$ from frame $t$ is determined. If $O_{j,k} \geq c_\mathit{over}$ or $j = \mathrm{argmax}_{i \in \hat{\mathcal K}_{x_{t}}} O_{i,k}$, the segments $k$ and $j$ are matched and receive the same id. If there is no match found for segment $k$ during this procedure, the quantity
\begin{equation}
    d = \min_{j \in \hat{\mathcal K}_{x_{t}}} \left\| \bar{j}_{t} - \bar{k}_{t-1} \right\|_{2} + \left\| \left( \bar{k}_{t-1} - \bar{k}_{t-2} \right) - \left( \bar{j}_{t} - \bar{k}_{t-1} \right)  \right\|_{2}
\end{equation}
is calculated for each available $j$ and both segments are matched if $d \leq c_\mathit{dist}$. This allows for matching segments that are closer to $\bar{k}_{t-1}$ than expected. If segment $k$ exists in frame $t-1$, but not in $t-2$, then step 2 is simplified: only the distance between the geometric center of $k \in \hat{\mathcal K}_{x_{t-1}}$ and $j \in \hat{\mathcal K}_{x_{t}}$ is computed and the segments are matched if the distance is smaller than $c_\mathit{dist}$.

\textbf{Step 3 (overlap).} $ \ $ If $t \geq 2$, the overlap $O_{j,k}$ of the segments $k \in \hat{\mathcal K}_{x_{t-1}}$ and $j \in \hat{\mathcal K}_{x_{t}}$ of two consecutive frames is calculated. If $O_{j,k} \geq c_\mathit{over}$ or $j = \mathrm{argmax}_{i \in \hat{\mathcal K}_{x_{t}}} O_{i,k}$, the segments $k$ and $j$ are matched. 

\textbf{Step 4 (regression).} $ \ $ In order to account for flashing predicted segments, either due to false prediction or occlusion, we implement a linear regression and match segments that are more than one, but at most $n_{lr}-2$, frames apart in temporal direction. 
If the id of segment $k \in \hat{\mathcal K}_{x_{*}}$, $* \in \{t-n_{lr}, \ldots, t-1\}$, in frame $t$ has not yet been assigned and $t \geq 4$, i.e., three frames have already been processed, then the geometric centers of segment $k$ are computed in frames $t-n_{lr}$ to $t-1$ (in case $k$ exists in all these frames). If at least two geometric centers are available, a linear regression is performed to predict the geometric center $(\hat{\bar{k}}_{t})$. If the distance between the predicted geometric center and the calculated geometric center of the segment $j \in \hat{\mathcal K}_{x_{t}}$ is less than a constant value $c_\mathit{lin}$, $k$ and $j$ are matched. If no match was found for segment $k$, segment $k \in \hat{\mathcal K}_{x_{t_{max}}}$ is shifted by the vector $\left( \hat{\bar{k}}_{t} - \bar{k}_{t_{max}} \right)$, where $t_{max} \in \{t-n_{lr}, \ldots, t-1\}$ denotes the frame where $k$ contains the maximum number of pixels. If $O_{j,k} \geq c_\mathit{over}$ or $j = \mathrm{argmax}_{i \in \hat{\mathcal K}_{x_{t}}} O_{i,k}$ applies to the resulting overlap, $k$ and $j$ are matched. 

\textbf{Step 5 (new ids).} $ \ $ All segments $j \in \hat{\mathcal K}_{x_{t}}$ that have not yet received an id are assigned with a new one. \\
%
%
\section{Segment-wise Metrics and Time Series}\label{sec:metrics}
In the previous chapter, we presented the semantic segmentation and the resulting probability distribution $f_{z}(y|x,w)$ for an image $x$, pixel $z$ and weights $w$. The degree of randomness in $f_{z}(y|x,w)$ is quantified by (pixel-wise) dispersion measures. Therefore we consider the entropy 
\begin{equation}
    E_z(x,w) =-\frac{1}{\log(c)}\sum_{y\in \mathcal{C}}f_z(y|x,w)\log f_z(y|x,w) \, ,
\end{equation}
the variation ratio 
\begin{equation}
    V_{z} = 1 - \max_{z}f_z(y|x,w)  
\end{equation}
and the probability margin $M_{z}$ ($V_{z}$ plus the second largest softmax output), as used in \cite{Schubert2019}.
A visualization of pixel-wise variation ratio is shown in \cref{fig:metrics}. Note that also other heat maps (like MC Dropout variance) can be processed. To obtain metrics per segment from these pixel-wise dispersion measures for each segment $k \in \hat{\mathcal K}_x$, we define mean dispersions $\bar D$ as
\begin{equation}
    \bar D = \frac{1}{S} \sum_{z\in k} D_z(x)
\end{equation}
where $D_{z} \in \{E_{z},V_{z},M_{z}\}$ and $S=|k|$ denotes the segment size. We further distinguish between the interior $k_{in}\subset k$, where a pixel $z \in k_{in}$ if all eight neighboring pixels of $z$ are elements of $k$, and the boundary $k_{bd}=  k \setminus k_{in}$. At segment level, the segment size and the mean disperions are provided as metrics, divided into interior and boundary. From these metrics, additional metrics are derived such as the relative segment sizes $\tilde S = S/S_{bd}$ and $\tilde S_{in} = S_{in}/S_{bd}$ as well as the relative mean dispersions $\tilde {\bar D} = \bar D \tilde S$ and $\tilde {\bar D}_{in} = \bar D_{in} \tilde S_{in}$ where $D \in \{E,V,M\}$ (see \cite{Rottmann2018,Schubert2019}). Our set of metrics is defined by these measures and the geometric center $\bar{k}$ (defined in (\ref{geom_center})) as well as the mean class probabilities for each class $y \in \{1,\ldots,c\}$
\begin{equation}
    P(y|k) = \frac{1}{S} \sum_{z \in k} f_{z}(y|x,w) \, .
\end{equation}
In summary, we use the following set of metrics:
\begin{align}
    U^{k} & = \ \{ \bar D, \bar D_{in}, \bar D_{bd}, \tilde {\bar D}, \tilde {\bar D}_{in} \, : \, D \in \{E,V,M\} \} \cup \{ \bar{k} \} \notag \\
    & \cup \{ S, S_{in}, S_{bd}, \tilde S, \tilde S_{in} \} \cup \{ P(y|k) \, : \, y=1,\ldots,c \} \, . 
\end{align}
The separate treatment of interior and boundary in all dispersion measures is motivated by typically large values of $D_z$ for $z\in k_{bd}$. In addition, we find that poor or false predictions are often accompanied by fractal segment shapes (which have a relatively large amount of boundary pixels, measurable by $\tilde S$ and $\tilde S_{in}$) and/or high dispersions $\bar D_{in}$ on the segment's interior.
The presented metrics are single-frame based, the proposed light-weight tracking method provides the identification of predicted segments in consecutive frames. Hence, we obtain time series for each of the defined metrics, that are subject to further analysis.
%
%
\section{Prediction of the IoU from Time Series}\label{sec:pred}
A measure to determine the prediction accuracy of the segmentation network with respect to the ground truth is the $\IoU$. In our test we use a slight modification, i.e., the adjusted $\IoU$ ($\sIoU$) which is less prone to fragmented objects, see \cite{Rottmann2018}.
In this work, we perform segment-wise predictions of the $\sIoU$ (meta regression) comparing different regression approaches and classify between $\sIoU = 0$ and $\sIoU > 0$ (meta classification), both for every predicted segment. Both prediction tasks are performed by means of the metrics introduced in \cref{sec:metrics}. Note that these metrics can be computed without the knowledge of the ground truth. Our aim is to analyze to which extent they are suitable for the meta tasks and how much we benefit from using time series.
For each segment $k \in \hat{\mathcal K}_{x_{t}}$ in frame $t$ we have the metrics $U^{k}_{t}$ and their history from previous frames due to the segment tracking. Both meta tasks are performed by means of the metrics $U^{k}_{i}$, $i = t-n_{c}, \ldots, t$, where $n_{c}$ describes the number of considered frames.
The methods used for meta classification are the logistic regression with $\ell_1$-penalty (LASSO \cite{Tibshirani1996}), gradient boosting regression \cite{Friedman2002} and a shallow neural network containing only a single hidden layer with $50$ neurons. The shorthand \textcolor{mediumorchid}{LR L1} refers to logistic regression, \textcolor{lightskyblue}{GB} to gradient boosting and \textcolor{plum(web)}{NN L2} to a neural network with $\ell_2$-penalty.
For meta regression we compare six different regression methods, this includes plain linear regression (\textcolor{cornflowerblue}{LR}), linear regression with $\ell_{1}$- and $\ell_{2}$-penalization (\textcolor{mediumorchid}{LR L1}/\textcolor{darkseagreen}{LR L2}), gradient boosting (\textcolor{lightskyblue}{GB}) and two shallow neural networks -- one with $\ell_{1}$-penalization and one with $\ell_{2}$ (\textcolor{mediumslateblue}{NN L1}/\textcolor{plum(web)}{NN L2}).
In addition to the presented time-dynamic components, we extend the approach from \cite{Rottmann2018} by incorporating further models for meta tasks, i.e., neural networks and gradient boosting. For this reason we consider linear models for meta tasks and single frame ($n_{c}=0$) based metrics as used in \cite{Rottmann2018} as baseline. Another approach presented in \cite{DeVries2018} also performs meta regression, however it is only designed for one object per image and on a single-frame basis. For this reason, we cannot regard this approach as a suitable baseline, a $\sim$150 (\#{}segments/image) CNN inferences per image are infeasible. 
%
%
%
\section{Numerical Results}\label{sec:result}
In this section, we investigate the properties of the metrics defined in the previous sections, the influence of the length of the time series considered and of different meta classification and regression methods. We perform our tests on two different datasets for the semantic segmentation of street scenes where also videos are available, the synthetic VIPER dataset \cite{Richter2017} obtained from the computer game GTA V and the KITTI dataset \cite{Geiger2013} with real street scene images from Karlsruhe, Germany. In all our tests we consider two different DeepLabv3+ networks \cite{Chen2018}. 
The DeepLabv3+ implementation and weights are available for two network backbones. First, there is the Xception65 network, a modified version of Xception \cite{Chollet2017} and it is a powerful structure for server-side deployment. On the other hand, is MobilenetV2 \cite{Sandler2018} a fast structure designed for mobile devices. Primarily we use Xception65 for VIPER and MobilenetV2 for KITTI, for the latter we also use Xception65 as a reference network to generate pseudo ground truth for the meta tasks. 
For tracking segments with our procedure, we assign the parameters defined in \cref{sec:tracking_segments} with the following values: $c_\mathit{near}=10$, $c_\mathit{over}=0.35$, $c_\mathit{dist}=100$ and $c_\mathit{lin}=50$. We study the predictive power of our $22$ metrics and segment-wise averaged class probabilities per segment and frame. From our tracking algorithm we get these metrics additionally from previous frames for every segment. 
\paragraph{VIPER Dataset.}
The VIPER dataset consists of more than $250$K $1,\!920 \times 1,\!080$ video frames and for all frames there is ground truth available, consisting of $23$ classes. We trained an Xception65 network starting from backbone weights for ImageNet \cite{Russakovsky2015}. 
We choose an output stride of $16$ and the input image is evaluated within the framework only on its original scale (DeepLabv3+ allows for evaluation on different scales and averaging the results). For a detailed explanation of the chosen parameters we refer to \cite{Chen2018}. We retrain the Xception65 network with $5,\!147$ training images and $847$ validation images (during training with different numbers of images we found that this number is sufficient). 
We only use images from the day category (i.e., bright images, no rain) for training and further processing, achieving a mean $\IoU$ of $50.33\%$. If we remove the classes mobile barrier, chair and van which are also underrepresented in the dataset (yielding $\IoU$s below $10\%$), the mean $\IoU$ rises to $57.38\%$. The DeepLabv3+ training pipeline does not include a test set as there is no ground truth available for the Cityscapes test set. The validation set is neither used for early stopping nor for parameter tuning. For meta classification and regression we use only $13$ video sequences consisting of $3,\!593$ images in total. From these images we obtain roughly $309,\!874$ segments (not yet matched over time) of which $251,\!368$ have non-empty interior. The latter are used in all numerical tests. We investigate the influence of time-dynamics on meta tasks, i.e., we firstly only present the segment-wise metrics $U^{k}_{t}$ of a single frame $t$ to the meta classifier/regressor, secondly we extend the metrics to time series with a length of up to $10$ previous time steps $U^{k}_{i}$, $i=t-10, \ldots, t-1$. We obtain $11$ different inputs for the meta tasks models. The presented results are averaged over $10$ runs obtained by random sampling of the train/validation/test splitting. In tables and figures, the corresponding standard deviations are given in brackets and by shades, respectively. Out of the $251,\!368$ segments with non-empty interior, $85,\!291$ have an $\sIoU=0$. 

First, we present results for meta classification, i.e., detection of false positive segments ($\sIoU=0$), using $38,\!000$ (randomly sampled) segments that are not presented to the segmentation network during training.  We apply a (meta) train/validation/test splitting of 70\%/10\%/20\% and evaluate the performance of different models for meta classification in terms of classification accuracy and AUROC. The AUROC is obtained by varying the decision threshold in a binary classification problem, here for the decision between $\sIoU=0$ and $>0$. 
We achieve test AUROC values of up to $86.01\% (\pm 0.56\%)$ and accuracies of up to $77.88\% (\pm 0.60\%)$. \Cref{tab:viper} shows the best results for different meta classification methods, i.e., logistic regression, a neural network and gradient boosting.
\begin{table}
\centering
\caption{Results for meta classification and regression for the different methods as well as for the naive and entropy baselines. The super script denotes the number of frames where the best performance and in particular the given values are reached. The best classification and regression results are highlighted.}
\label{tab:viper}
\scalebox{0.72}{
\begin{tabular}{||c||c|c|c||}
\cline{1-4}
\multicolumn{4}{||c||}{Meta Classification $\sIoU=0,>0$} \\
\cline{1-4}
\multicolumn{2}{||c}{Naive Baseline:} & \multicolumn{1}{c}{ACC = $66.07\%$} & \multicolumn{1}{c||}{AUROC = $50.00\%$ } \\
\cline{1-4}
\multicolumn{2}{||c}{Entropy Baseline:} & \multicolumn{1}{c}{ ACC = $68.43\%(\pm0.29\%)$} & \multicolumn{1}{c||}{AUROC = $74.02\%(\pm0.32\%)$ } \\
\cline{1-4}
\multicolumn{1}{||c}{ } & $\textbf{\textcolor{mediumorchid}{LR L1}}$  &  $\textbf{\textcolor{lightskyblue}{GB}}$ &$\textbf{\textcolor{plum(web)}{NN L2}}$ \\
\cline{1-4}
ACC & $75.75\%(\pm0.49\%)^8$ & $\mathbf{77.88}\boldsymbol{\%}(\pm0.60\%)^2$ & $76.62\%(\pm0.51\%)^6$ \rule{0mm}{3.0mm} \\
\cline{1-4}
AUROC & $83.44\%(\pm0.47\%)^7$ & $\mathbf{86.01}\boldsymbol{\%}(\pm0.56\%)^4$ & $84.52\%(\pm0.50\%)^{11}$ \rule{0mm}{3.0mm} \\
\cline{1-4}
\multicolumn{4}{||c||}{Meta Regression $\sIoU$} \\ 
\cline{1-4}
\multicolumn{2}{||c}{Entropy Baseline:} &  \multicolumn{1}{c}{ $\sigma$ = $0.178(\pm0.000)$} & \multicolumn{1}{c||}{$R^2$ = $64.18\%(\pm0.34\%)$ \rule{0mm}{3.0mm} } \\
\cline{1-4}
\multicolumn{1}{||c}{ } & $\textbf{\textcolor{cornflowerblue}{LR}}$  &  $\textbf{\textcolor{mediumorchid}{LR L1}}$ &$\textbf{\textcolor{darkseagreen}{LR L2}}$ \\
\cline{1-4}
$\sigma$ & $0.124(\pm0.002)^6$ & $0.124(\pm0.002)^7$ & $0.124(\pm0.002)^5$ \rule{0mm}{3.0mm} \\
\cline{1-4}
$R^2$ & $82.58\%(\pm0.45\%)^6$ & $82.56\%(\pm0.43\%)^7$ & $82.57\%(\pm0.44\%)^5$ \rule{0mm}{3.0mm} \\
\cline{1-4}
\multicolumn{1}{||c}{ } &$\textbf{\textcolor{lightskyblue}{GB}}$  &  $\textbf{\textcolor{mediumslateblue}{NN L1}}$ &$\textbf{\textcolor{plum(web)}{NN L2}}$ \\
\cline{1-4}
$\sigma$ & $\mathbf{0.112}(\pm0.002)^6$ & $0.118(\pm0.002)^4$ & $0.117(\pm0.002)^2$ \rule{0mm}{3.0mm} \\ 
\cline{1-4}
$R^2$ & $\mathbf{85.82}\boldsymbol{\%}(\pm0.36\%)^6$ & $84.36\%(\pm0.51\%)^4$ & $84.58\%(\pm0.44\%)^2$ \rule{0mm}{3.0mm} \\ 
\cline{1-4}
\end{tabular} }
\end{table}
The super script denotes the number of frames where the best performance and in particular the given values are reached. We observe that the best results are achieved when considering more than one frame. Furthermore, significant differences between the methods for the meta tasks
can be observed, gradient boosting shows the best performance with respect to accuracy and AUROC.

\begin{figure}[t]
    \subfigure{\includegraphics[width=0.24\textwidth]{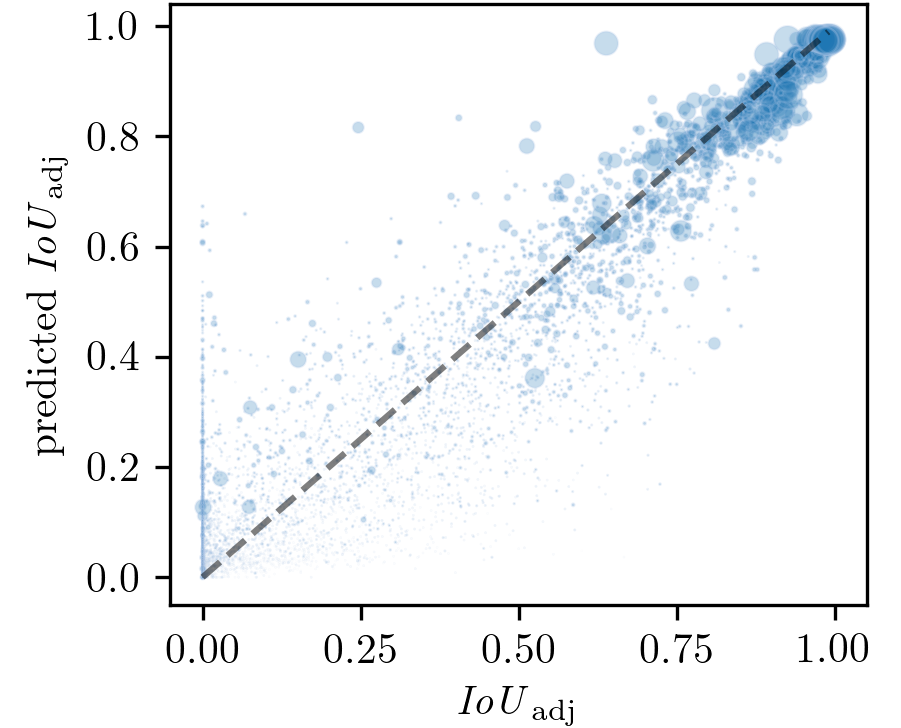}}
    \subfigure{\includegraphics[width=0.24\textwidth]{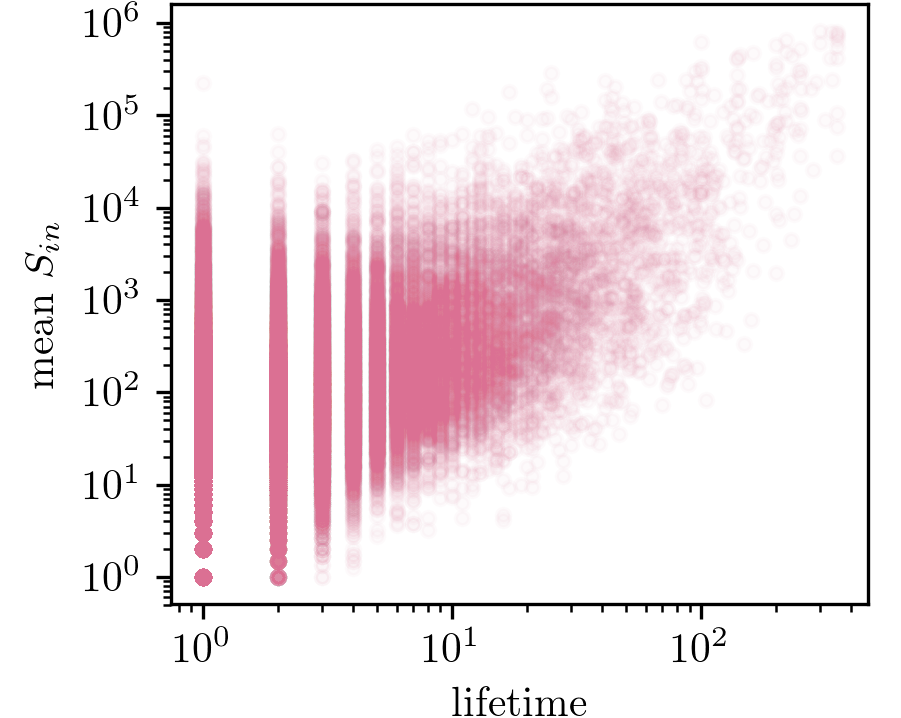}}
    \caption{Predicted $\sIoU$ vs.\ $\sIoU$ for all non-empty segments (\emph{left}). The dot size is proportional to the segment size. Segment lifetime (time series length) vs.\ mean interior segment size, both on log scale (\emph{right}).}
    \label{fig:iou_vs}
\end{figure}

\begin{figure}[t]
    \center
    \includegraphics[width=0.47\textwidth]{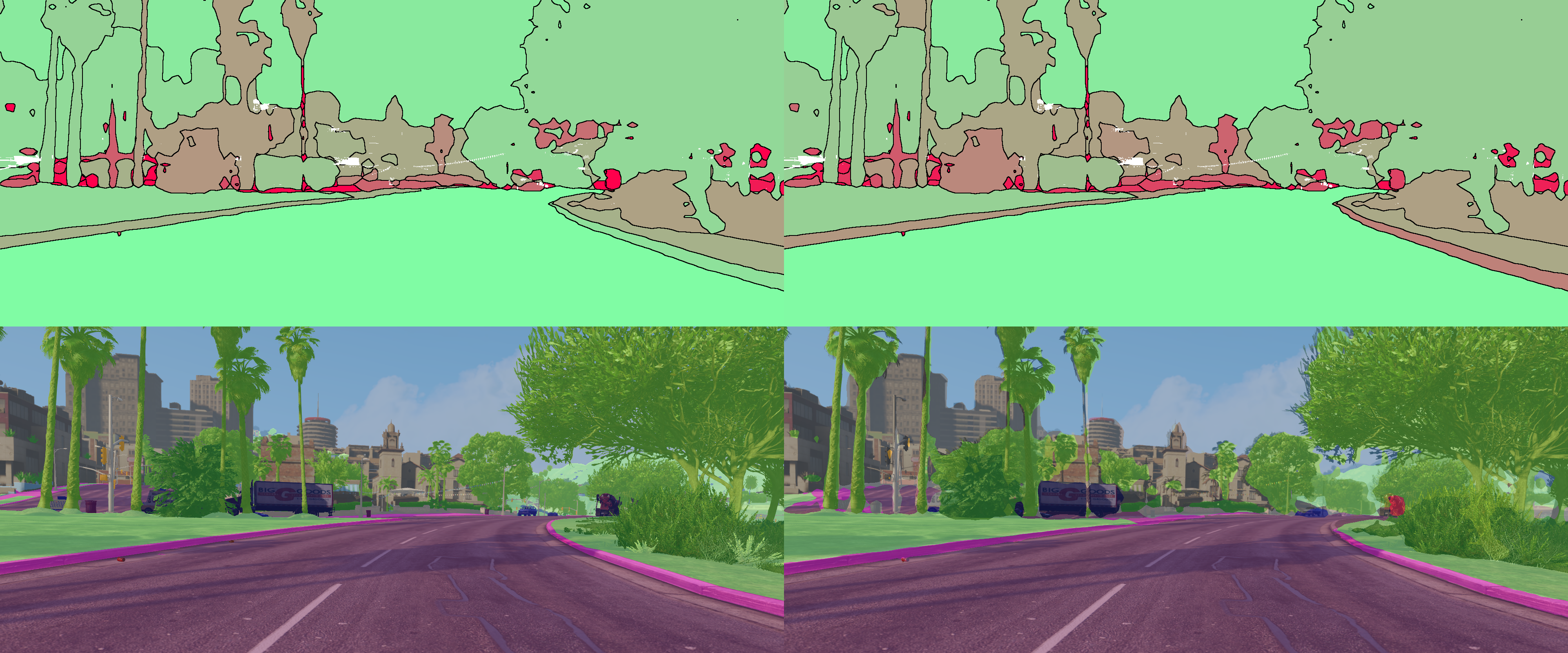}
    \caption{Ground truth image (\emph{bottom left}), prediction obtained by a neural network (\emph{bottom right}), a visualization of the true segment-wise $\sIoU$ of prediction and ground truth (\emph{top left}) and its prediction obtained from meta regression (\emph{top right}). Green color corresponds to high $\sIoU$ values and red color to low ones. For the white regions there is no ground truth available, these regions are not included in the statistical evaluation.}
    \label{fig:pred_iou_gta}
\end{figure}

Next, we predict $\sIoU$ values via meta regression to estimate prediction quality. For this task we state regression standard errors $\sigma$ and $R^2$ values. We achieve $R^{2}$ values of up to $85.82\% (\pm 0.36\%)$. This value is obtained by gradient boosting incorporating $5$ previous frames. For this particular study, the correlation of the calculated and predicted $\sIoU$ is depicted in \cref{fig:iou_vs} (left), an illustration of the resulting quality estimate is given in \cref{fig:pred_iou_gta}. 
We also provide video sequences that visualize the $\sIoU$ prediction and the segment tracking, 
see \url{https://youtu.be/TQaV5ONCV-Y}.
Results for meta regression are also summarized in \cref{tab:viper}, the findings are in analogy to those for meta classification. Gradient boosting performs best, and more frames yield better results than a single one. The approach in \cite{Rottmann2018} can be considered as a baseline for both meta tasks, since we extend this single-frame based method by time series and further classification/regression models. 
The results in \cite{Rottmann2018} were compared with the mean entropy per segment as a single-metric baseline and with a naive baseline which is given by random guessing (randomly assigning a probability to each segment and then tresholding on it). The classification accuracy is the number of correct predictions divided by the total number of predictions made. It is maximized for the threshold being either $1$ if we have more $\sIoU > 0$ than $\sIoU = 0$ segments, or $0$ else. The corresponding AUROC value is $50\%$. For the entropy baseline we use single-frame gradient boosting. \Cref{tab:viper} includes these two baselines, both are clearly outperformed.
\Cref{fig:iou_vs} (right) depicts the correlation of the time series length and the mean interior segment size. On average, a predicted segment exists for $4.4$ frames, however when we consider only segments that contain at least 1,000 interior pixels, the average life time increases to $19.9$ frames.
\paragraph{KITTI Dataset.} 
For the KITTI dataset, we use both DeepLabv3+ networks (pre-trained on the Cityscapes dataset \cite{Cordts2016}, available on GitHub), however, for the evaluation we primarily use MobilenetV2. As parameters for the Xception65 network we choose an output stride of $8$, a decoder output stride of $4$ and an evaluation of the input on scales of $0.75$, $1.00$ and $1.25$ (averaging the results). For the MobilenetV2 we use an output stride of $16$ and the input image is evaluated within the framework only on its original scale. 
In our tests we use $29$ street scene videos consisting of $12,\!223$ images with a resolution of $1,\!392 \times 512$. Of these images, only $142$ are labelled. An evaluation of the meta tasks requires a train/val/test splitting. Therefore, the small number of labeled images seems almost insufficient. Hence, we acquire alternative sources of useful information besides the (real) ground truth. First, we utilize the Xception65 network with high predictive performance, its predicted segmentations we term \emph{pseudo ground truth}. We generate pseudo ground truth for all images where ground truth is not available. The mean $\IoU$ performance of the Xception65 network for the 142 labelled images is $64.54\%$, for the MobilenetV2 the mean $\IoU$ is $50.48\%$. In addition, to augment the structured dataset of metrics, we apply a variant of SMOTE for continuous target variables for data augmentation \cite{Torgo2013}.
This is particularly important when only working with the scarce real ground truth, in our tests we observed that SMOTE prevents overfitting in this scenario.
The train/val/test splitting of the data with ground truth available is the same as for the VIPER dataset, i.e., 70\%/10\%/20\%. The shorthand ``augmented'' refers to data obtained from SMOTE, ``pseudo'' refers to pseudo ground truth obtained from the Xception65 network and ``real'' refers to ground truth obtained from a human annotator. These additions are only used during training. We utilize the Xception65 network only for the generation of pseudo ground truth, all tests are performed using the MobilenetV2. The KITTI dataset consists of $19$ classes ($4$ classes less than VIPER).

\begin{figure} 
    \center
    \includegraphics[width=0.49\textwidth]{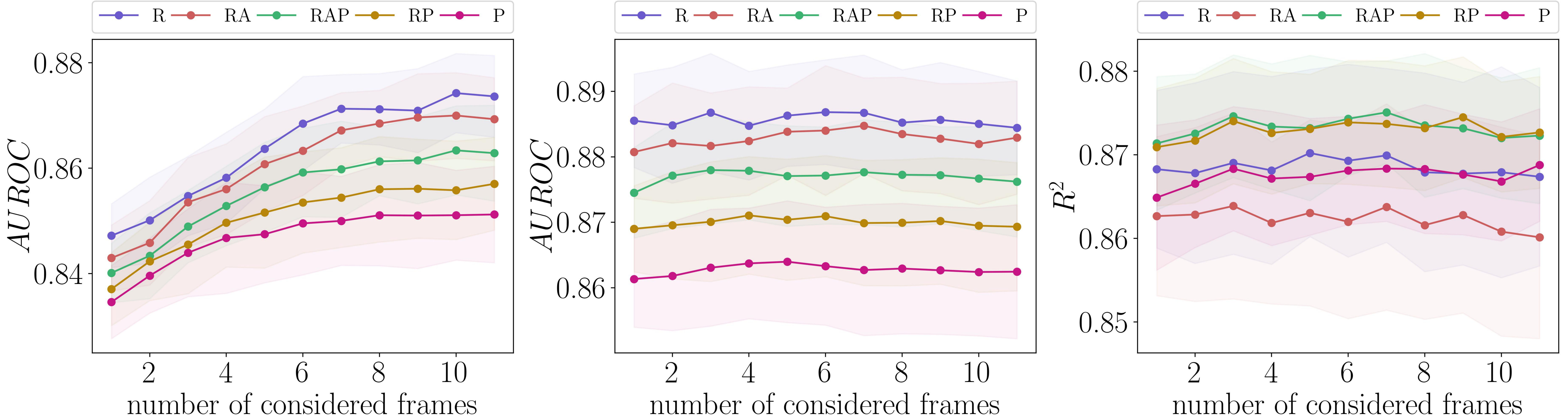}
\caption{A selection of results for meta classification AUROC and regression $R^2$ as functions of the number of frames and for different compositions of training data (see \cref{tab:train_val_test}). Meta classification via a neural network with $\ell_{2}$-penalization (\emph{left}), via gradient boosting (\emph{middle}) as well as meta regression via gradient boosting (\emph{right}).}
\label{fig:class_timeline}
\end{figure}

From the $12,\!223$ chosen images, we obtain $452,\!287$ segments of which $378,\!984$ have non-empty interior. Of these segments, $129,\!033$ have an $\sIoU=0$. A selection of results for meta classification AUROC and regression $R^2$ as functions of the number of frames, i.e., the maximum time series length, is given in \cref{fig:class_timeline}. 
The approach of \cite{Rottmann2018} corresponds to the single-frame results in these plots.
The meta classification results for neural networks presented in \cref{fig:class_timeline} (left) indeed show, that an increasing length of time series has a positive effect on meta classification. 
On the other hand, the results in \cref{fig:class_timeline} (middle) show that gradient boosting does not benefit as much from time series. In both cases augmentation and pseudo ground truth do not improve the models' performance on the test set and although the neural network benefits a lot from time series, its best performance is still about $1\%$ below that of gradient boosting. With respect to the influence of time series length, the results for meta regression with gradient boosting in \cref{fig:class_timeline} (right) are qualitatively similar to those in \cref{fig:class_timeline} (middle). However, we observe in this case that the incorporation of pseudo ground truth slightly increases the performance. Noteworthily, for the $R^2$ values we achieve the best results with the training set consisting of real, augmented and pseudo ground truth in two out of six models (see \cref{tab:kitti}), demonstrating that SMOTE helps in this case. 
\begin{table}[t]
\centering
\caption{Results for meta classification and regression for different compositions of training data and methods. 
The dataset for the entropy baseline is selected such that the baseline performance is maximized. The super script denotes the number of frames where the best performance and thus the given value is reached. The best results for each data composition are highlighted.}
\label{tab:kitti}
\scalebox{0.735}{
\begin{tabular}{||c||c||c|c|c||}
\cline{1-5}
\multicolumn{5}{||c||}{Meta Classification $\sIoU=0,>0$} \\
\cline{1-5}
\multicolumn{5}{||c||}{Naive Baseline: \ \ \ \ \ ACC = $65.95\%$ \ \ \ \ \ AUROC = $50.00\%$ } \\
\cline{1-5}
\multicolumn{5}{||c||}{Entropy Baseline for \textbf{\textcolor{mediumred-violet}{P}}: \ \ \ \ \ ACC = $68.66\%(\pm1.82\%)$ \ \ \ \ \ AUROC = $75.81\%(\pm1.68\%)$ } \\
\cline{1-5}
\multicolumn{2}{||c}{ } & $\textbf{\textcolor{mediumorchid}{LR L1}}$  &  $\textbf{\textcolor{lightskyblue}{GB}}$ & $\textbf{\textcolor{plum(web)}{NN L2}}$ \\
\cline{1-5}
\multicolumn{1}{||c||}{ } & \textbf{\textcolor{slateblue}{R}} & $\mathbf{76.69}\boldsymbol{\%}(\pm1.68\%)^{10}$ & $\mathbf{81.20}\boldsymbol{\%}(\pm1.02\%)^4$ & $\mathbf{79.67}\boldsymbol{\%}(\pm0.93\%)^{10}$ \rule{0mm}{3.0mm}    \\ 
\multicolumn{1}{||c||}{ } & \textbf{\textcolor{indianred}{RA}} & $76.60\%(\pm1.31\%)^7$ & $80.73\%(\pm1.03\%)^9$ & $78.62\%(\pm0.61\%)^{11}$    \\ 
ACC & \textbf{\textcolor{mediumseagreen}{RAP}} & $76.18\%(\pm1.22\%)^7$ & $79.64\%(\pm1.03\%)^7$ & $77.08\%(\pm1.05\%)^9$   \\
\multicolumn{1}{||c||}{ } & \textbf{\textcolor{darkgoldenrod}{RP}} & $76.52\%(\pm0.80\%)^8$ & $78.45\%(\pm0.88\%)^8$ & $76.35\%(\pm0.67\%)^9$    \\
\multicolumn{1}{||c||}{ } & \textbf{\textcolor{mediumred-violet}{P}} & $75.96\%(\pm0.80\%)^{11}$ & $77.56\%(\pm0.95\%)^5$ & $75.68\%(\pm0.67\%)^{11}$   \\
\cline{1-5}
\multicolumn{1}{||c||}{ } & \textbf{\textcolor{slateblue}{R}} & $85.13\%(\pm0.84\%)^1$ & $\mathbf{88.68}\boldsymbol{\%}(\pm0.80\%)^6$ & $\mathbf{87.42}\boldsymbol{\%}(\pm0.75\%)^{10}$ \rule{0mm}{3.0mm}    \\ 
\multicolumn{1}{||c||}{ } & \textbf{\textcolor{indianred}{RA}} & $85.00\%(\pm1.05\%)^7$ & $88.47\%(\pm0.73\%)^7$ & $87.00\%(\pm0.81\%)^{10}$  \\ 
AUROC & \textbf{\textcolor{mediumseagreen}{RAP}} & $\mathbf{85.39}\boldsymbol{\%}(\pm0.97\%)^6$ & $87.80\%(\pm0.82\%)^3$ & $86.34\%(\pm0.84\%)^{10}$ \\ 
\multicolumn{1}{||c||}{ } & \textbf{\textcolor{darkgoldenrod}{RP}} & $85.38\%(\pm0.87\%)^8$ & $87.11\%(\pm0.90\%)^4$ & $85.70\%(\pm0.88\%)^{11}$   \\ 
\multicolumn{1}{||c||}{ } & \textbf{\textcolor{mediumred-violet}{P}} & $84.94\%(\pm1.03\%)^6$ & $86.40\%(\pm0.93\%)^5$ & $85.12\%(\pm0.92\%)^{11}$   \\ 
\cline{1-5}
\multicolumn{5}{||c||}{Meta Regression $\sIoU$} \\ 
\cline{1-5}
\multicolumn{5}{||c||}{Entropy Baseline for \textbf{\textcolor{darkgoldenrod}{RP}}: \ \ \ \ \ $\sigma$ = $0.167(\pm0.006)$ \ \ \ \ \ $R^2$ = $71.05\%(\pm2.58\%)$ \rule{0mm}{3.0mm} }  \\
\cline{1-5}
\multicolumn{2}{||c}{ } & $\textbf{\textcolor{cornflowerblue}{LR}}$  &  $\textbf{\textcolor{mediumorchid}{LR L1}}$ &$\textbf{\textcolor{darkseagreen}{LR L2}}$ \\
\cline{1-5}
\multicolumn{1}{||c||}{ } & \textbf{\textcolor{slateblue}{R}} & $\mathbf{0.128}(\pm0.003)^2$ & $0.129(\pm0.003)^2$ & $\mathbf{0.128}(\pm0.003)^2$ \rule{0mm}{3.0mm}    \\ 
\multicolumn{1}{||c||}{ } & \textbf{\textcolor{indianred}{RA}} & $0.134(\pm0.003)^2$ & $0.134(\pm0.003)^3$ & $0.134(\pm0.003)^2$    \\
$\sigma$ & \textbf{\textcolor{mediumseagreen}{RAP}} & $0.129(\pm0.003)^7$ & $0.129(\pm0.003)^7$ & $0.129(\pm0.003)^7$   \\ 
\multicolumn{1}{||c||}{ } & \textbf{\textcolor{darkgoldenrod}{RP}} & $\mathbf{0.128}(\pm0.003)^7$ & $\mathbf{0.128}(\pm0.002)^7$ & $\mathbf{0.128}(\pm0.003)^7$   \\
\multicolumn{1}{||c||}{ } & \textbf{\textcolor{mediumred-violet}{P}} & $\mathbf{0.128}(\pm0.003)^7$ & $0.129(\pm0.002)^7$ & $0.129(\pm0.003)^7$    \\
\cline{1-5}
\multicolumn{1}{||c||}{ } & \textbf{\textcolor{slateblue}{R}} & $83.48\%(\pm0.99\%)^2$ & $83.37\%(\pm0.92\%)^2$ & $83.49\%(\pm0.96\%)^2$ \rule{0mm}{3.0mm}    \\ 
\multicolumn{1}{||c||}{ } & \textbf{\textcolor{indianred}{RA}} & $82.06\%(\pm0.96\%)^2$ & $82.09\%(\pm0.94\%)^3$ & $82.08\%(\pm0.95\%)^2$   \\ 
$R^2$ & \textbf{\textcolor{mediumseagreen}{RAP}} & $83.38\%(\pm0.89\%)^7$ & $83.35\%(\pm0.90\%)^7$ & $83.40\%(\pm0.92\%)^7$    \\ 
\multicolumn{1}{||c||}{ } & \textbf{\textcolor{darkgoldenrod}{RP}} & $\mathbf{83.62}\boldsymbol{\%}(\pm0.91\%)^7$ & $\mathbf{83.54}\boldsymbol{\%}(\pm0.88\%)^7$ & $\mathbf{83.61}\boldsymbol{\%}(\pm0.91\%)^7$    \\ 
\multicolumn{1}{||c||}{ } & \textbf{\textcolor{mediumred-violet}{P}} & $83.43\%(\pm0.90\%)^7$ & $83.36\%(\pm0.86\%)^7$ & $83.41\%(\pm0.91\%)^7$  \\ 
\cline{1-5}
\multicolumn{2}{||c}{ } & $\textbf{\textcolor{lightskyblue}{GB}}$  & $\textbf{\textcolor{mediumslateblue}{NN L1}}$ &$\textbf{\textcolor{plum(web)}{NN L2}}$ \\
\cline{1-5}
\multicolumn{1}{||c||}{ } & \textbf{\textcolor{slateblue}{R}} & $0.114(\pm0.004)^5$ & $\mathbf{0.114}(\pm0.005)^1$ & $\mathbf{0.113}(\pm0.005)^1$ \rule{0mm}{3.0mm}    \\
\multicolumn{1}{||c||}{ } & \textbf{\textcolor{indianred}{RA}} & $0.116(\pm0.004)^3$ & $0.118(\pm0.007)^1$ & $0.116(\pm0.005)^1$   \\
$\sigma$ & \textbf{\textcolor{mediumseagreen}{RAP}} & $\mathbf{0.112}(\pm0.003)^7$ & $\mathbf{0.114}(\pm0.003)^1$ & $0.114(\pm0.005)^1$     \\
\multicolumn{1}{||c||}{ } & \textbf{\textcolor{darkgoldenrod}{RP}} & $\mathbf{0.112}(\pm0.002)^9$ & $0.116(\pm0.004)^1$ & $0.115(\pm0.003)^2$    \\
\multicolumn{1}{||c||}{ } & \textbf{\textcolor{mediumred-violet}{P}} & $0.114(\pm0.002)^{11}$ & $0.118(\pm0.004)^1$ & $0.117(\pm0.004)^3$   \\
\cline{1-5}
\multicolumn{1}{||c||}{ } & \textbf{\textcolor{slateblue}{R}} & $87.02\%(\pm1.00\%)^5$ & $86.98\%(\pm1.07\%)^1$ & $\mathbf{87.16}\boldsymbol{\%}(\pm1.25\%)^1$ \rule{0mm}{3.0mm}    \\ 
\multicolumn{1}{||c||}{ } & \textbf{\textcolor{indianred}{RA}} & $86.39\%(\pm1.11\%)^3$ & $85.94\%(\pm1.76\%)^1$ & $86.46\%(\pm1.32\%)^1$   \\ 
$R^2$ & \textbf{\textcolor{mediumseagreen}{RAP}} & $\mathbf{87.51}\boldsymbol{\%}(\pm0.61\%)^7$ & $\mathbf{87.03}\boldsymbol{\%}(\pm0.71\%)^1$ & $86.97\%(\pm1.10\%)^1$   \\ 
\multicolumn{1}{||c||}{ } & \textbf{\textcolor{darkgoldenrod}{RP}} & $87.45\%(\pm0.72\%)^9$ & $86.51\%(\pm0.88\%)^1$ & $86.69\%(\pm0.85\%)^2$    \\ 
\multicolumn{1}{||c||}{ } & \textbf{\textcolor{mediumred-violet}{P}} & $86.88\%(\pm0.67\%)^{11}$ & $86.13\%(\pm0.95\%)^1$ & $86.24\%(\pm0.99\%)^3$   \\ 
\cline{1-5}
\end{tabular} }
\end{table}
An overview of the different compositions of training data and the train/val/test splitting are given in \cref{tab:train_val_test}. 
\begin{table}[t]
\centering
\caption{Train/val/test splitting, different compositions of training data and their approximate number of segments.}
\label{tab:train_val_test}
\scalebox{0.8}{
\begin{tabular}{||l||l|l||c||}
\cline{1-4}
splitting & \multicolumn{2}{c||}{types of data / annotation}  & no. of segments \\
\cline{1-4}
      & \textbf{\textcolor{slateblue}{R}}        & real                       & $\sim$ 3,400 \\
      & \textbf{\textcolor{indianred}{RA}}       & real and augmented         & $\sim$ 27,000 \\
train & \textbf{\textcolor{mediumseagreen}{RAP}} & real, augmented and pseudo & $\sim$ 27,000 \\
      & \textbf{\textcolor{darkgoldenrod}{RP}}   & real and pseudo            & $\sim$ 27,000 \\
      & \textbf{\textcolor{mediumred-violet}{P}} & pseudo                     & $\sim$ 27,000 \\
\cline{1-4}                                                                         
val   &                                          & real                       & $\sim$ 500 \\
\cline{1-4}
test  &                                          & real                       & $\sim$ 1,000 \\
\cline{1-4}
\end{tabular} }
\end{table}
Furthermore, gradient boosting trained with real ground truth and gradient boosting trained only with pseudo ground truth perform almost equally well. This shows that meta regression can be learned when there is no ground truth but a strong reference model available. Note that this (except for the data augmentation part) is in accordance to our findings for the VIPER dataset. Results for a wider range of tests (including those previously discussed) are summarized in \cref{tab:kitti}. Again we provide video sequences, 
see \url{https://youtu.be/YcQ-i9cHjLk}.
For meta tasks, we achieve accuracies of up to $81.20\% (\pm 1.02\%)$ and AUROC values of up to $88.68\% (\pm 0.80\%)$ as well as $R^{2}$ values of up to $87.51\% (\pm 0.61\%)$. 
As for the VIPER dataset, we outperform the analogous baselines (see \cref{tab:kitti}). 
As the labeled 142 images only yield 4,877 segments, we observe overfitting in our tests for all models when increasing the length of the time series. This might serve as an explanation that in some cases, time series do not increase performance. In particular, we observe overfitting in our tests when using gradient boosting, this holds for both datasets, KITTI and VIPER. It is indeed well-known that gradient boosting requires plenty of data.
%
\section{Conclusion and Outlook}
In this work we extended the approach presented in \cite{Rottmann2018} by incorporating time series as input for meta classification and regression. To this end, we introduced a light-weight tracking algorithm for semantic segmentation. From matched segments we generated time series of metrics and used these as inputs for the meta tasks. In our tests we studied the influence of the time series length on different models for the meta tasks, i.e., gradient boosting, neural networks and linear ones. Our results show significant improvements in comparison to those presented in \cite{Rottmann2018}. More precisely, in contrast to the single-frame approach using only linear models, we increase the accuracy by $6.78$ pp and the AUROC by $5.04$ pp. The $R^2$ value for meta regression 
is increased by $5.63$ pp. As a further improvement, we plan to develop additional time-dynamic metrics, as the presented metrics are still single-frame based. 
The source code of our method is publicly available at \url{https://github.com/kmaag/Time-Dynamic-Prediction-Reliability}.

\bibliographystyle{IEEEtran}
\bibliography{biblio}

\end{document}